\documentclass[10pt, conference, letterpaper]{IEEEtran}
\usepackage{cite}
\usepackage[caption=false,font=footnotesize]{subfig}
\usepackage{url}
\usepackage[utf8]{inputenc} % this is needed for umlauts
\usepackage[USenglish,UKenglish,english]{babel} % this is needed for umlauts
\usepackage[T1]{fontenc}    % this is needed for correct output of umlauts in pdf
\makeatletter
\newcommand{\upquotetype}{}
\newcommand{\upquote@aux}[1]{\text{\upquotetype}#1\text{\upquotetype}}
\newcommand{\upquotesingle}{\renewcommand{\upquotetype}{\textquotesingle}\upquote@aux}
\newcommand{\upquotedouble}{\renewcommand{\upquotetype}{\textquotedbl}\upquote@aux}

\usepackage{amsthm}

\usepackage{textcomp}

\usepackage{graphicx} % more modern
\usepackage{epstopdf}
\usepackage{amsfonts}
\usepackage{amsmath}

\usepackage{algorithm}
\usepackage{algorithmic}

\begin{document}
\title{Topic Grids for Homogeneous Data Visualization}

\author{\IEEEauthorblockN{Shih-Chieh Su\IEEEauthorrefmark{1},
Joseph Vaughn\IEEEauthorrefmark{2} and Jean-Laurent Huynh\IEEEauthorrefmark{3}}
\IEEEauthorblockA{Information Security and Risk Management Department\\
Qualcomm Inc.\\
San Diego, CA, 92121\\
Email: \IEEEauthorrefmark{1}shihchie@qualcomm.com,
\IEEEauthorrefmark{2}jmvaughn@qualcomm.com,
\IEEEauthorrefmark{3}jeanlaur@qualcomm.com}}

\maketitle

\begin{abstract}
We propose the topic grids to detect anomaly and analyze the behavior based on the access log content. Content-based behavioral risk is quantified in the high dimensional space where the topics are generated from the log. The topics are being projected homogeneously into a space that is perception- and interaction-friendly to the human experts.
\end{abstract}

\IEEEpeerreviewmaketitle

\section{Introduction}\label{sec-intro}

To make the data points in the high dimensional space $\mathcal{H}$ visible to human, a word embedding (or dimension reduction) technique is employed to map the data points to a lower dimensional space $\mathcal{L}$. The word embedding technique of choice attempts to preserve some relationship among the data points in $\mathcal{H}$ after mapping them to $\mathcal{L}$. 

The output from the dimension reduction algorithm is a set of data points that are non-uniformly scattered around the visualization space. This helps to explain the clustering behavior, including inter-cluster and intra-cluster, among the data points. However, there are also some drawbacks:
\begin{enumerate}
  \item The clusters tangle with others; some data points overlap with others. Overlap makes the information less perceivable.
  \item The data points are denser in some area. The heterogeneity makes human interaction with the data points more difficult.
\end{enumerate}

In order to better utilize the visualization space, we propose to distribute the data points evenly over the visualization space. The cloud of data points is deformed in the same space defined by the dimension reduction algorithm of choice.

\section{Method}

The algorithm we propose is called the split-diffuse (SD) algorithm (Algorithm~\ref{alg-SD}). It attempts to realize the strategy above.

\begin{algorithm}[tb]
   \caption{Split-diffuse algorithm (square of power of $2$)}
   \label{alg-SD}
   {\bfseries Input:} data points $\{p\}$ of length $2^h \times 2^h$, depth $d=0$, allocation string $c=\upquotesingle{ }$\\ 
   {\bfseries split-diffuse ($\{p\}$, $d$, $c$)}
\begin{algorithmic}
   \STATE $k \leftarrow $ length of $\{p\}$
   \IF{$k = 1$,} 
   \STATE resolve $\mathbb{S}(p)$ from $c$
   \STATE return $p$
   \ENDIF
   \STATE $a \leftarrow mod($depth$,2)$
   \STATE $m \leftarrow $ median of $\{p\}$ in the dimension $a$
   \STATE 
   \begin{tabular}{@{}ll@{}l} 
   return & ( & [{\bfseries split-diffuse} ($\{p:p\leq m|_{dim=a}\}$, $d$+1, $c$+\upquotesingle{L})], \\ 
   & & [{\bfseries split-diffuse} ($\{p:p> m|_{dim=a}\}$, $d$+1, $c$+\upquotesingle{R})])
   \end{tabular}
\end{algorithmic}
\end{algorithm}

We keep track of the splitting path in string $c$. At the end of the recursion, the placement each single point $p$ is resolved. The indexes of the SD-mapped points, $\mathbb{S}(p)$, are all integers, and forms a $2^h \times 2^h$ array. This means that the mapped data points are equally spaced in a $2^h \times 2^h$ square. To achieve this uniformity in the space $\mathcal{L}$, the data points are essentially diffused from the denser area to the coarser area by the SD algorithm --- hence the name split-diffuse.

\begin{figure*}[htp]
  \centering
  \subfloat[Sample MDS output]{\includegraphics[scale=0.23]{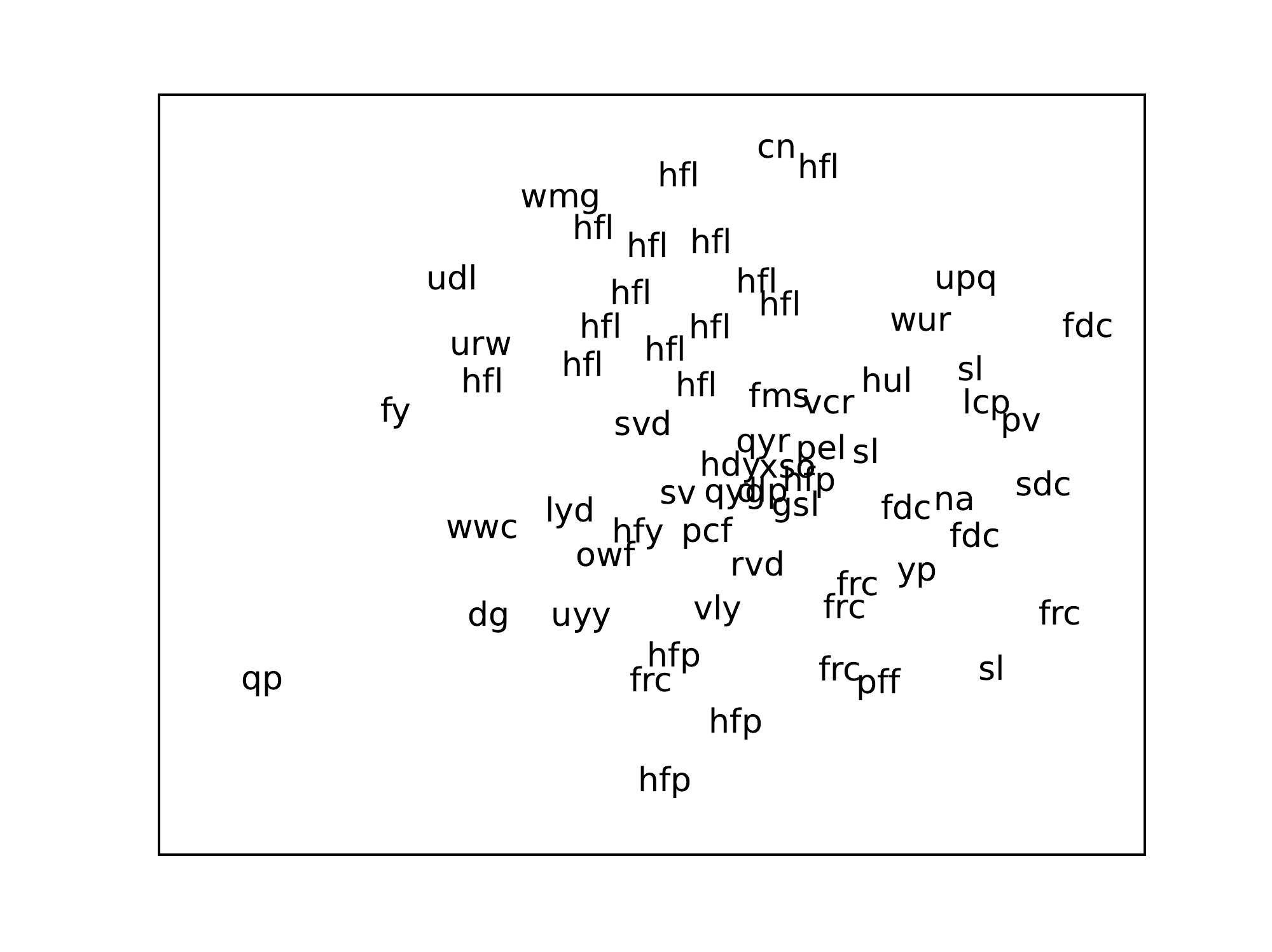}}
  \subfloat[Sample SD output from (a)]{\includegraphics[scale=0.23]{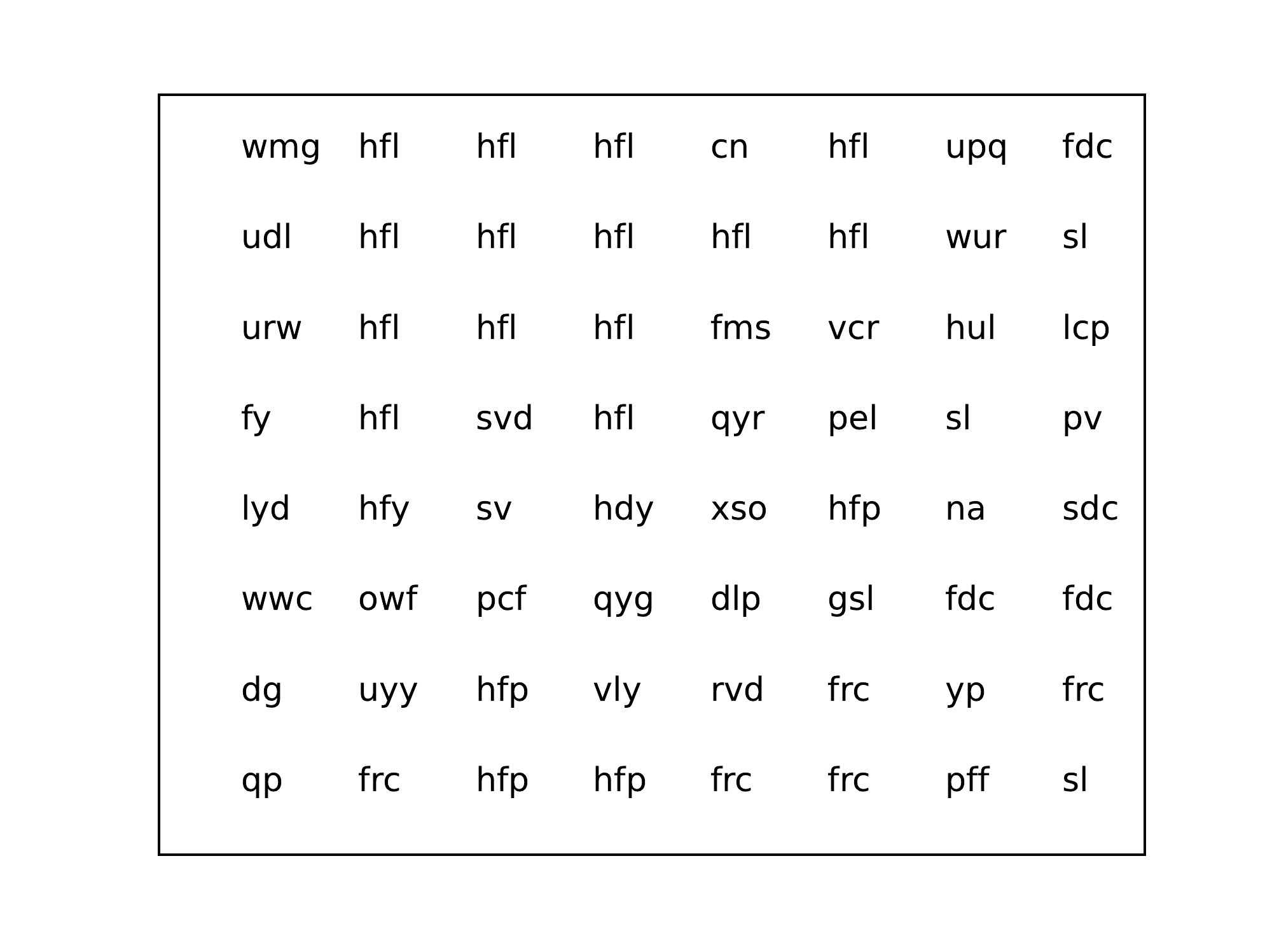}}
  \subfloat[Sample t-SNE output]{\includegraphics[scale=0.23]{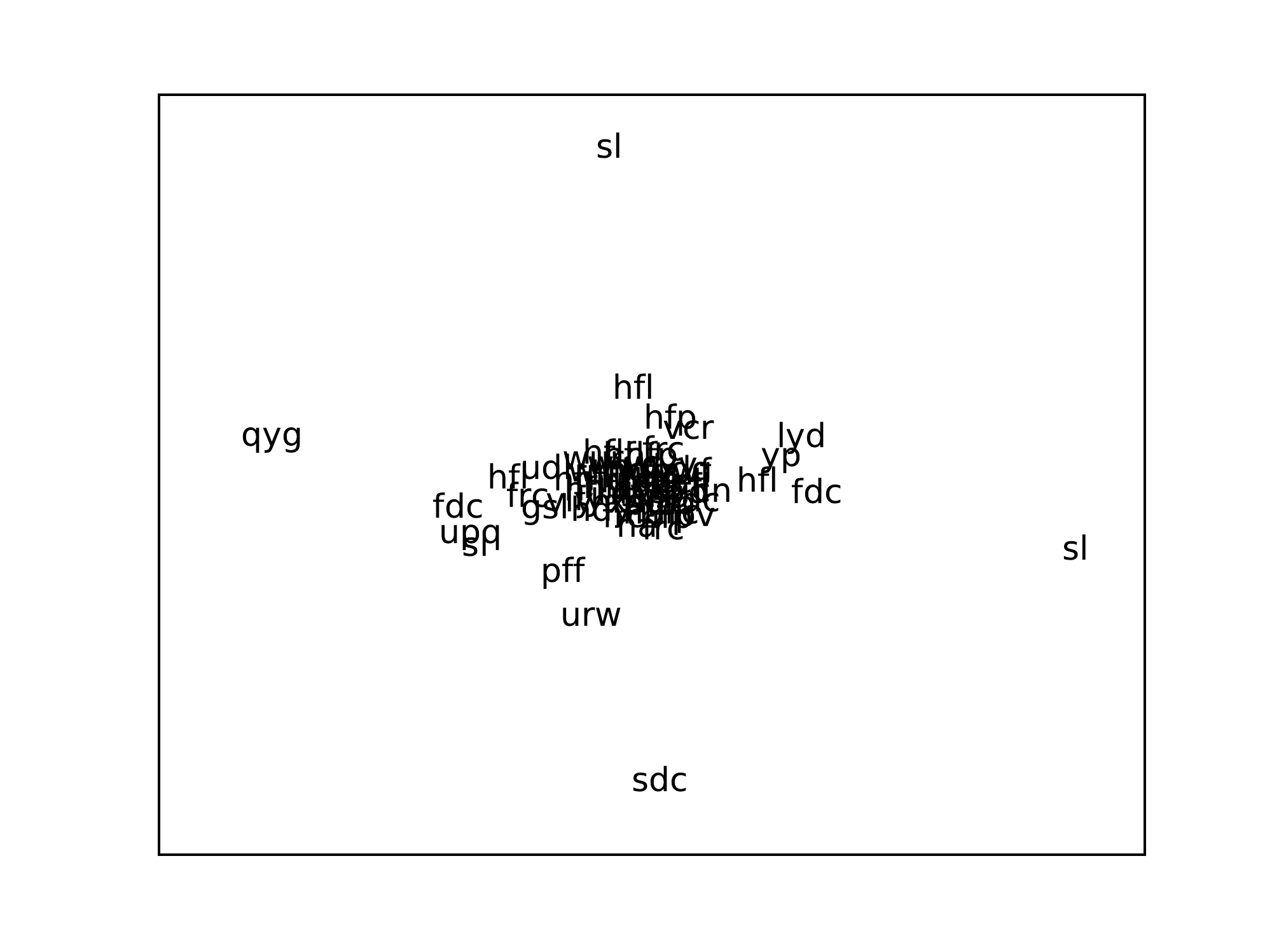}}
  \subfloat[Sample SD output from (c)]{\includegraphics[scale=0.23]{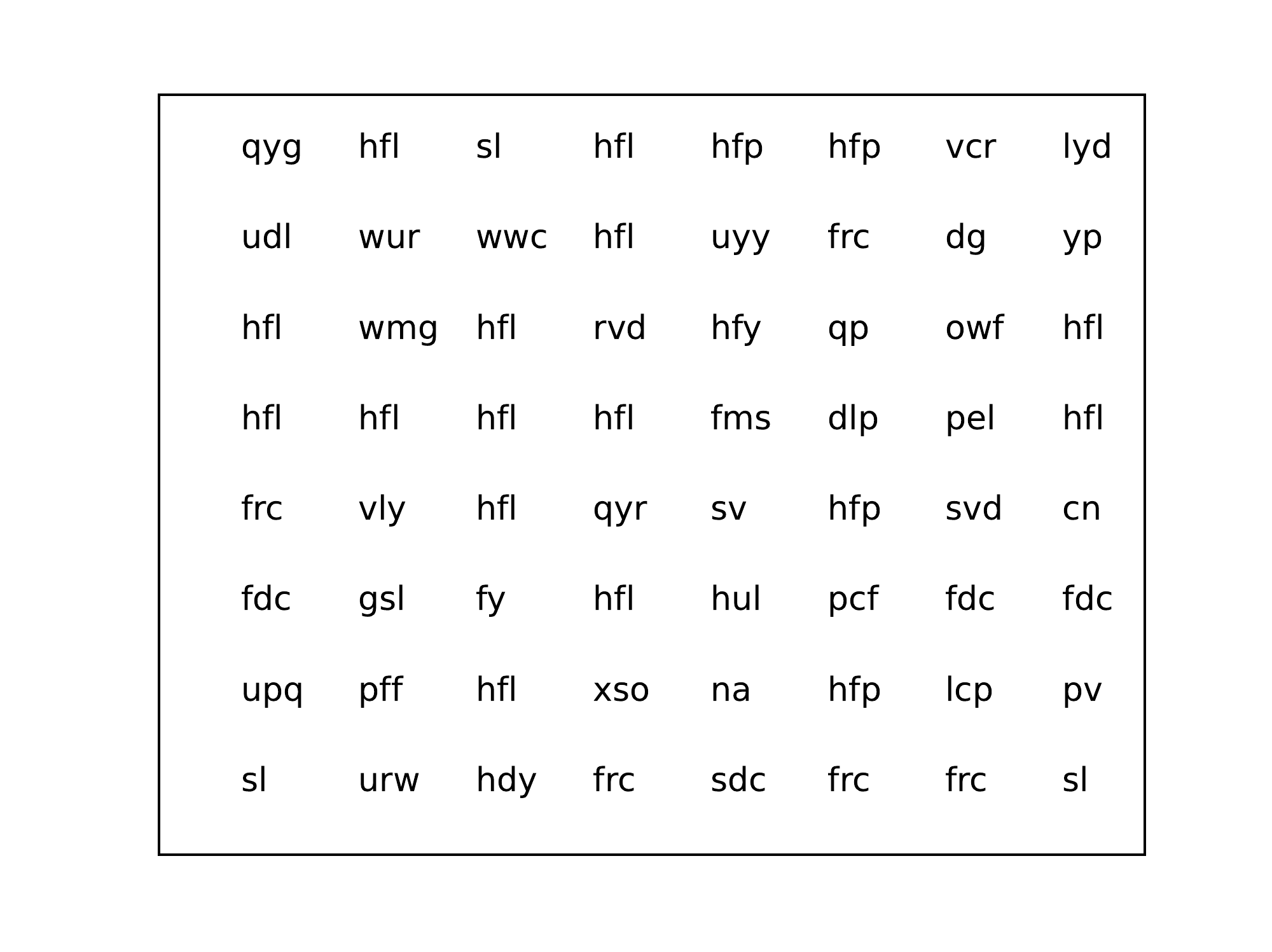}}
\caption{Examples of placing $64$ data points over an $8\times8$ layout with the SD algorithm.}
\label{fig-mdstsne}
\end{figure*}

\begin{figure*}[htp]
  \centering
  \subfloat[current activities of a user]{\includegraphics[scale=0.64]{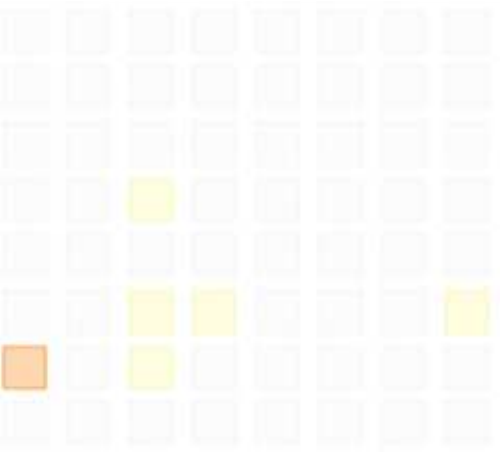}}\quad
  \subfloat[historical activities of this user]{\includegraphics[scale=0.64]{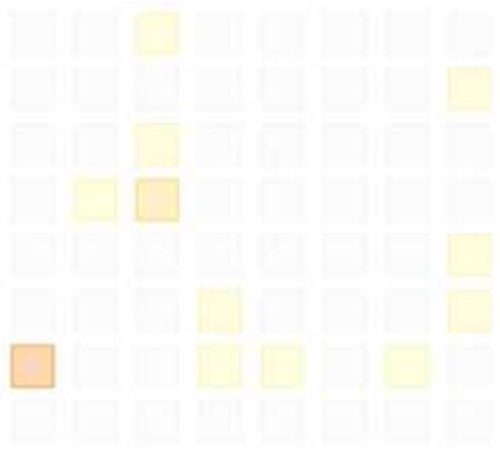}}\quad
  \subfloat[risk against historical self]{\includegraphics[scale=0.64]{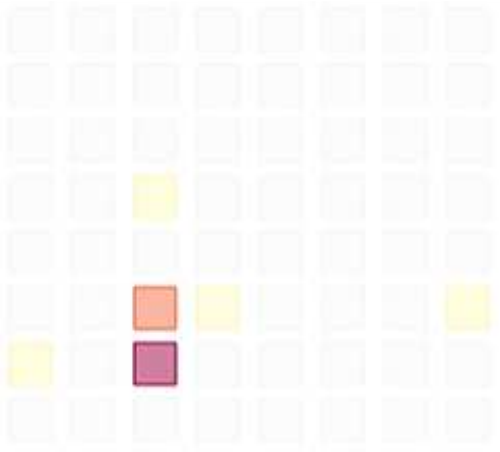}}\quad
  \subfloat[historical activities of the peers of this user]{\includegraphics[scale=0.64]{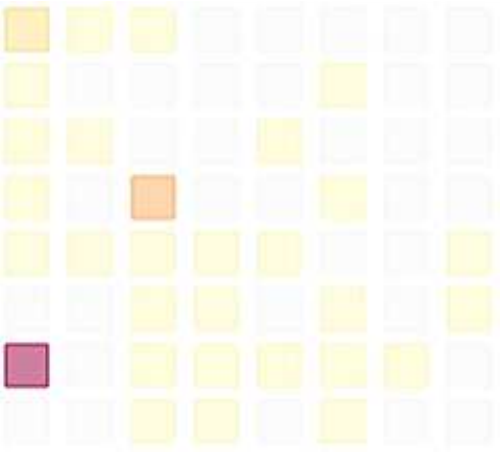}}\quad
  \subfloat[risk against peers]{\includegraphics[scale=0.64]{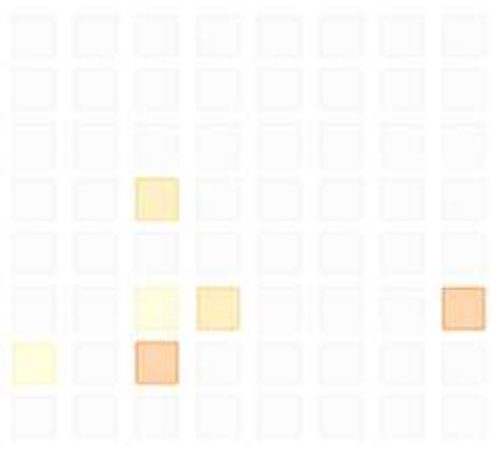}}
\caption{The topic grids. The self risk in (c) is derived from comparing the current activities (a) and the historical activities (b) of a specific user. The peer risk in (e) is derived from comparing the current activities (a) and the peers' activities (d) of a specific user.}
\label{fig-TG}
\end{figure*}

In Figure~\ref{fig-mdstsne}, we generate $64$ topics regarding to the content of some repository access logs. A topic is represented by the first three letters of the most descriptive word. The three letters are then encrypted. There are topics sharing the same most descriptive word, but they are different topics, and can further be distinguished by the less descriptive words. 

In the high dimensional space $\mathcal{H}$ (in our case $\mathcal{H}$ has 19K+ dimensions) where the topics are created, the topics sharing the same representative word tend to stay close to each other. The $64$ topics are mapped from $\mathcal{H}$ to a 2D space $\mathcal{L}$, via MDS \cite{Torgerson52} and t-SNE \cite{Van08}. The SD algorithm follows to produce the uniform placement. Having the topics sharing the same representative word stay close to each other after the SD mapping is another desirable feature.
 
\section{Performance} 

\begin{table}[t]
\caption{Mean error ratio over various layouts.}
\label{table-mer}
\begin{center}
\begin{small}
\begin{sc}
\begin{tabular}{ccccc}
\hline
Layout & Sampling & Constraints & $Err_{I}$ & $Err_{II}$ \\
\hline
$4\times4$ & $\mathbb{U}$   & 240   & 0.2042 & 0.0292 \\
$8\times8$ & $\mathbb{U}$   & 4,032  & 0.1347 & 0.0270 \\
$16\times16$ & $\mathbb{U}$ & 65,280 & 0.0776 & 0.0192 \\
$32\times32$ & $\mathbb{U}$ & 1,047,552 & 0.0426 & 0.0124 \\
$64\times64$ & $\mathbb{U}$ & 16,773,120 & 0.0228 & 0.0074 \\
\hline
$4\times4$ & $\mathbb{G}$   & 240   & 0.2368 & 0.0618 \\
$8\times8$ & $\mathbb{G}$   & 4,032  & 0.1845 & 0.0769 \\
$16\times16$ & $\mathbb{G}$ & 65,280 & 0.1459 & 0.0875 \\
$32\times32$ & $\mathbb{G}$ & 1,047,552 & 0.1242 & 0.0940 \\
$64\times64$ & $\mathbb{G}$ & 16,773,120 & 0.1131 & 0.0977 \\
\hline
\end{tabular}
\end{sc}
\end{small}
\end{center}
\end{table}

In order to measure the performance of the SD algorithm, we randomly generate the topics in the 2D space $\mathcal{L}$. We define $Err_{I}$ to be the ratio of violated topology constraints. For example, when $p_i$ is to the right of $p_j$, the placement of $\mathbb{S}(p_i)$ not being to the right of $\mathbb{S}(p_j)$ is a violation. There are totally $\|\{p\}\|\cdot(\|\{p\}\|-1)$ such constraints in a 2D space $\mathcal{L}$. A loosen metric $Err_{II}$ is defined so that the constraint in dimension $a$ is not violated when $\mathbb{S}(p_i)|_a=\mathbb{S}(p_j)|_a$.

The random topics are generated over a 2D space via the uniform approach $\mathbb{U}$, where topics are sampled within a square; and the Gaussian approach $\mathbb{G}$, where topics are sampled from two-variate Gaussian distribution with the magnitude in one variant doubling the other, and then rotated by $\pi/4$.

$Err_{I}$ decreases as the grid set size increases on both $\mathbb{U}$ and $\mathbb{G}$ samplings. However, $Err_{II}$ only decreases as the grid set size increases on the $\mathbb{U}$ sampling. $Err_{II}$ increases as the grid set size increases on the $\mathbb{G}$ sampling. Moreover, although $Err_{I}$ decreases as the grid set size increases on both $\mathbb{U}$ and $\mathbb{G}$ samplings, the decrease rate varies. On the $\mathbb{U}$ sampling, $Err_{I}$ decreases almost $90\%$ when the layout increase from $4\times4$ to $64\times64$. However, $Err_{I}$ only decreases around half of its value during the same layout increase for the $\mathbb{G}$ sampling.

\section{Application and Conclusion}

We apply the SD algorithm to help analyzing behavioral content in the cyber security domain. The goal of the system is to detect behavioral anomaly based on the access logs. A log entry usually has a timestamp, a unique identifier of a user, an action, and the content on which the action was performed. After proper punctuation, the path, the content, and/or any meta data regarding to this access can be viewed as a document, called the content document. We use the latent Dirichlet allocation (LDA) model \cite{Blei03} to decide the topics among all these content documents over a benchmark period of time.

In our system, $64$ topics are generated at a word vector space of $19K+$ dimensions. The relevance between a content document and each individual topic is measured.  The anomaly, or risk, of an access is quantified by comparing the topics involved in this access with the topics involved in the historical accesses of the same user, as well as the topics involved in the accesses of the peers of this user.

The output of the SD algorithm for our system is a set of topic grids, where the $64$ topics are placed over a $8\times8$ layout. The same set of topic grids is used to render the amount of activities on each topic and the risk on each topic. One use case is presented in Figure~\ref{fig-TG}. The human expert can interact with the grids and get detailed explanation about the topic (e.g., mouse over for topic content and click for access detail). Such type of interaction is hard to achieve with the conventional word embedding techniques. like the ones in Figure~\ref{fig-mdstsne} (a) and (c).

In addition to the cyber security domain, the topic grids can be applied to other domains like e-commerce, credit card transaction, customer service, wherever the content document can be derived from the log files.

%\section*{Acknowledgment}

%The authors would like to thank Qualcomm for sponsoring the development of the topic grids.

\bibliographystyle{IEEEtran}
\bibliography{IEEEabrv,jsu2016}

\begin{thebibliography}{1}
\providecommand{\url}[1]{#1}
\csname url@samestyle\endcsname
\providecommand{\newblock}{\relax}
\providecommand{\bibinfo}[2]{#2}
\providecommand{\BIBentrySTDinterwordspacing}{\spaceskip=0pt\relax}
\providecommand{\BIBentryALTinterwordstretchfactor}{4}
\providecommand{\BIBentryALTinterwordspacing}{\spaceskip=\fontdimen2\font plus
\BIBentryALTinterwordstretchfactor\fontdimen3\font minus
  \fontdimen4\font\relax}
\providecommand{\BIBforeignlanguage}[2]{{%
\expandafter\ifx\csname l@#1\endcsname\relax
\typeout{** WARNING: IEEEtran.bst: No hyphenation pattern has been}%
\typeout{** loaded for the language `#1'. Using the pattern for}%
\typeout{** the default language instead.}%
\else
\language=\csname l@#1\endcsname
\fi
#2}}
\providecommand{\BIBdecl}{\relax}
\BIBdecl

\bibitem{Torgerson52}
W.~S. Torgerson, ``Multidimensional scaling: I. theory and method,''
  \emph{Psychometrika}, vol.~17, no.~4, pp. 401--419, 1952.

\bibitem{Van08}
L.~Van~der Maaten and G.~E. Hinton, ``Visualizing data using {t-SNE},''
  \emph{Journal of Machine Learning Research}, vol.~9, no. 2579-2605, p.~85,
  2008.

\bibitem{Blei03}
D.~M. Blei, A.~Y. Ng, and M.~I. Jordan, ``Latent dirichlet allocation,''
  \emph{Journal of Machine Learning Research}, vol.~3, pp. 993--1022, 2003.

\end{thebibliography}

\end{document}